\pdfoutput=1
\documentclass[11pt]{article}

\usepackage[preprint]{acl}

\usepackage{times}
\usepackage{latexsym}
\usepackage{amsmath}
\usepackage{amssymb}
\usepackage{dblfloatfix}

\usepackage[T1]{fontenc}
\usepackage[utf8]{inputenc}

\usepackage{microtype}

\usepackage{inconsolata}

\usepackage{graphicx}

\usepackage{hyperref}

\title{Citegeist: Automated Generation of Related Work Analysis on the arXiv Corpus}

\author{
  Claas Beger\textsuperscript{*}\thanks{Equal contribution.} \\
  Cornell University \\
  \texttt{cbb89@cornell.edu} \\
  \And
  Carl-Leander Henneking\textsuperscript{*} \\
  Cornell University \\
  \texttt{ch2273@cornell.edu}
}
\begin{document}
\maketitle
\begin{abstract}
Large Language Models provide significant new opportunities for the generation of high-quality written works. However, their employment in the research community is inhibited by their tendency to hallucinate invalid sources and lack of direct access to a knowledge base of relevant scientific articles. In this work, we present Citegeist: An application pipeline using dynamic Retrieval Augmented Generation (RAG) on the arXiv Corpus to generate a related work section and other citation-backed outputs. For this purpose, we employ a mixture of embedding-based similarity matching, summarization, and multi-stage filtering. To adapt to the continuous growth of the document base, we also present an optimized way of incorporating new and modified papers. To enable easy utilization in the scientific community, we release both, a \href{https://citegeist.org}{website}, as well as an implementation harness that works with several different LLM implementations.
\end{abstract}

\section{Introduction}
\label{sec:Motivation}
Significant advances in language generation quality in recent years have enabled the employment of Large Language Models (LLMs) as writing assistants for various fields and purposes. However, while the quality of produced texts is often high, LLMs face critical challenges in domains requiring factual precision and verifiable citations, such as academic writing. Their tendency to hallucinate invalid or non-existent sources \cite{info:doi/10.2196/53164, magesh2024hallucinationfreeassessingreliabilityleading, RePEc:sae:amerec:v:69:y:2024:i:1:p:80-87}, combined with the lack of direct integration with up-to-date scientific knowledge bases, limits their applicability in research contexts. To address the former challenge, \cite{lewis2021retrievalaugmentedgenerationknowledgeintensivenlp} introduced Retrieval-Augmented-Generation (RAG) by directly employing a Wikipedia-based vector index to insert external knowledge into the model context. 

This work aims to transition this approach to the arXiv corpus, a popular and ever-growing dataset consisting of approximately 2.6 million academic papers. The vast size, frequent updates, and the availability of a dedicated Python library make the arXiv an intuitive choice for research-centered retrieval-augmented generation (RAG). However, this scale imposes challenges in data interaction, as the full-text dataset exceeds 1 TB. To mitigate this, we perform multiple filtering steps based on abstracts and retrieve full documents only for identified relevant sources. We then construct a vector database of abstract embeddings to enable efficient similarity search.

Additionally, recognizing that the related work requirements vary across research fields and paper contents, we introduce three key hyperparameters—\textit{breadth}, \textit{depth}, and \textit{diversity}—which are employed in our retrieval algorithm. We also place a strong emphasis on proper citation techniques and rigorously evaluate our approach in terms of source relevancy and the writing quality of the generated results.

Beyond the generation of related work sections, we adapt our method to answer scientific questions with citation-based responses, yielding promising results.\newline

\noindent \textbf{Key Contributions:}
\begin{itemize}
    \setlength{\itemsep}{0pt} % Reduces space between items
    \item Development of a dynamic retrieval and synthesis application for related work generation.
    \item Introduction of three key hyperparameters—\textit{breadth}, \textit{depth}, and \textit{diversity}—to fine-tune the content and style of the result.
    \item Support for uploading full PDFs to enhance content-based retrieval.
    \item Employment of full paper texts through alternating between importance weighting and summarization techniques.
\end{itemize}

\begin{figure*}[htb]
    \centering
    \includegraphics[width=0.95\textwidth]{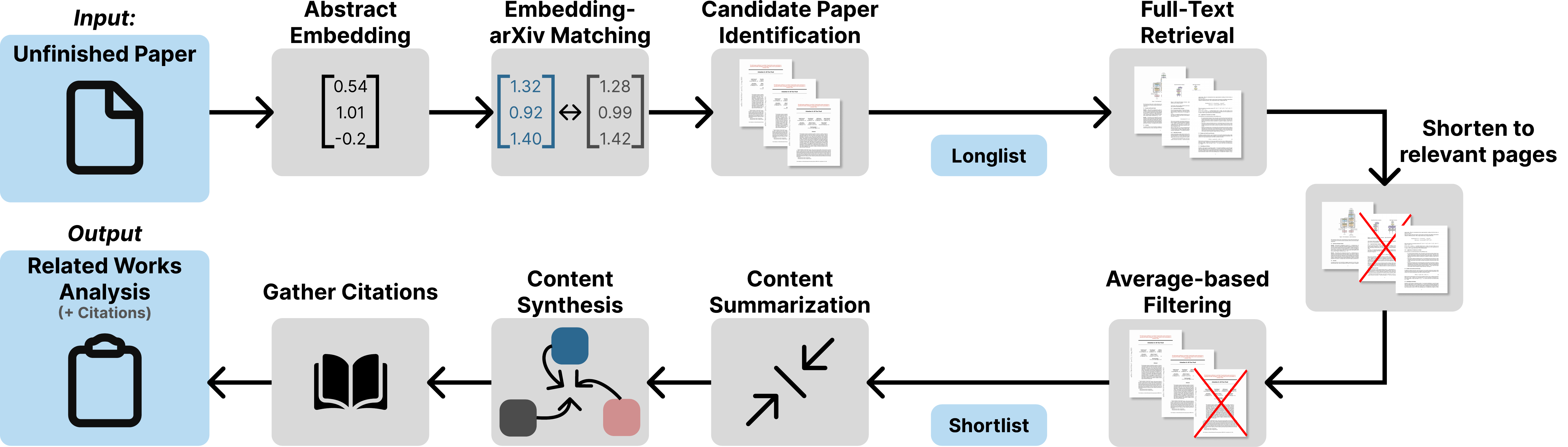}
    \caption{Our pipeline takes either an abstract or a full paper as input and produces a related works analysis, incl. all citations. We initially retrieve a set of candidate works that we filter in multiple steps, and then we summarize and synthesize relevant content to produce the related works analysis.}
    \label{fig:pipeline}
\end{figure*}

\section{Related Work}
\label{sec:related-work}
Recent advancements using RAG have been shown to enhance performance on knowledge-intensive tasks by integrating external sources \cite{lewis2021retrievalaugmentedgenerationknowledgeintensivenlp}. Building on this, \citet{jiang2023activeretrievalaugmentedgeneration} and \citet{trivedi2023interleavingretrievalchainofthoughtreasoning} demonstrate how retrieval datasets can be dynamically updated to incorporate the latest knowledge. Alternatively, \citet{yao2023reactsynergizingreasoningacting} demonstrates how data may be accessed directly through Agentic Workflow. Our project aims to use the arXiv corpus \cite{arxiv_org_submitters_2024} for a dynamic RAG system applied to the task of related work analysis. These works are highly relevant to our approach since arXiv experiences close to 20.000 article submissions a month (see \autoref{fig:arXiv_Growth}) and correct data migration is non-trivial. This issue is especially prominent for applications that explicitly train a model for information or document retrieval \cite{pmlr-v202-kishore23a}. Furthermore, there has been research on the efficiency of such retrieval models for scientific articles \cite{ajith2024litsearchretrievalbenchmarkscientific} with a focus on extractive question-answering but without utilization of summarization or RAG. 

In recent literature, related work generation has emerged as a prominent task within the domain of scientific multi-document summarization \cite{li2022automaticrelatedworkgeneration}. However, many prior approaches rely heavily on a predefined citation list \cite{inproceedings, Chen2019AutomaticGO, 8931592, Deng2021AutomaticRW}. In contrast, we propose an independent extraction pipeline that does not require citation data, addressing a significant gap in the existing body of work. Our method leverages multi-stage retrieval, filtering, and summarization techniques to enhance related work generation. To the best of our knowledge, few works have explored this direction. For instance, \citet{githubGitHubAllenaiai2scholarqalib} focuses on synthesizing large bodies of academic literature, but their approach is designed for general literature reviews rather than for extracting related work tailored to a specific input abstract. Similarly, \citet{asai2024openscholarsynthesizingscientificliterature} aims at unifying relevant scientific papers for a given query, with a primary focus on scientific question answering. However, their method does not incorporate hyperparameters to fine-tune results, nor does it support processing full PDFs. Furthermore, our approach distinguishes itself by employing a dynamic combination of summarization and relevance estimation during full-text processing.
\section{Method}
\label{sec:method}

We leverage an existing arXiv metadata corpus, which contains abstracts and paper IDs. To build up our vector database, we extract abstracts and perform an embedding operation using the all-mpnet-base-v2 \cite{henderson2019repositoryconversationaldatasets} Sentence Transformer. We chose this model because of its applicability to paragraphs of up to about 384-word pieces, which is suitable for academic abstracts and filtered page contents. Using this model, we embed all abstracts contained in the corpus and generate a corresponding hash, which we utilize later to refresh the existing database efficiently. Further, we compute a topic assignment using a fine-tuned BERTopic version \cite{grootendorst2022bertopic}, which can be used to produce content-based subsets. Based on this data, we build up our vector database using Milvus \cite{10.1145/3448016.3457550}, which is optimized for similarity-based lookups.

\subsection{Retrieval and Generation}
\label{subsec:retrieval-and-generation}
Our proposed architecture works on an excerpt of the target paper, most of the time this will be the abstract (alternatives are discussed later in this section). We embed this abstract and perform a similarity-based search using a cosine-similarity metric on our embedding metadata. Based on the results, we extract candidate abstracts. We parameterize the selection criteria to account for different user requirements. The selection starts with the abstract that has the highest similarity score. Thereafter, the next paper \( i^* \) is chosen iteratively according to
\begin{center}
\vspace{-0.5cm}
\begingroup
\small
\begin{equation} 
\label{eq:weighted-similarity}
    i^* = \arg\max_{i \notin S} \left[ (1 - w) \cdot s_i + 
    w \cdot \left( 1 - \min_{p_j \in S} \operatorname{sim}(e_i, e_j) \right) \right]
\end{equation}
\endgroup
\end{center}

Where \( P = \{p_1, p_2, \dots, p_n\} \) is the set of all papers in the query result, \( S \) is the subset of chosen papers, and each paper \( p_i \) has an abstract embedding \( e_i \in \mathbb{R}^d \). The parameter \( w \in [0, 1] \) determines the trade-off between prioritizing similarity and ensuring diversity. \( \operatorname{sim}(e_i, e_j) \) is the cosine similarity, and \( s_i \) denotes similarity to the input abstract. Generally, we define \( w \) as the \textit{diversity} of the generation. We also consider the \textit{breadth}, which affects the size of the initial paper set, as well as the number of selected candidates. We refer to the result of this process as the \textit{longlist} of candidate papers.

\begin{figure*}[h!]
    \centering
    \includegraphics[width=0.8\linewidth]{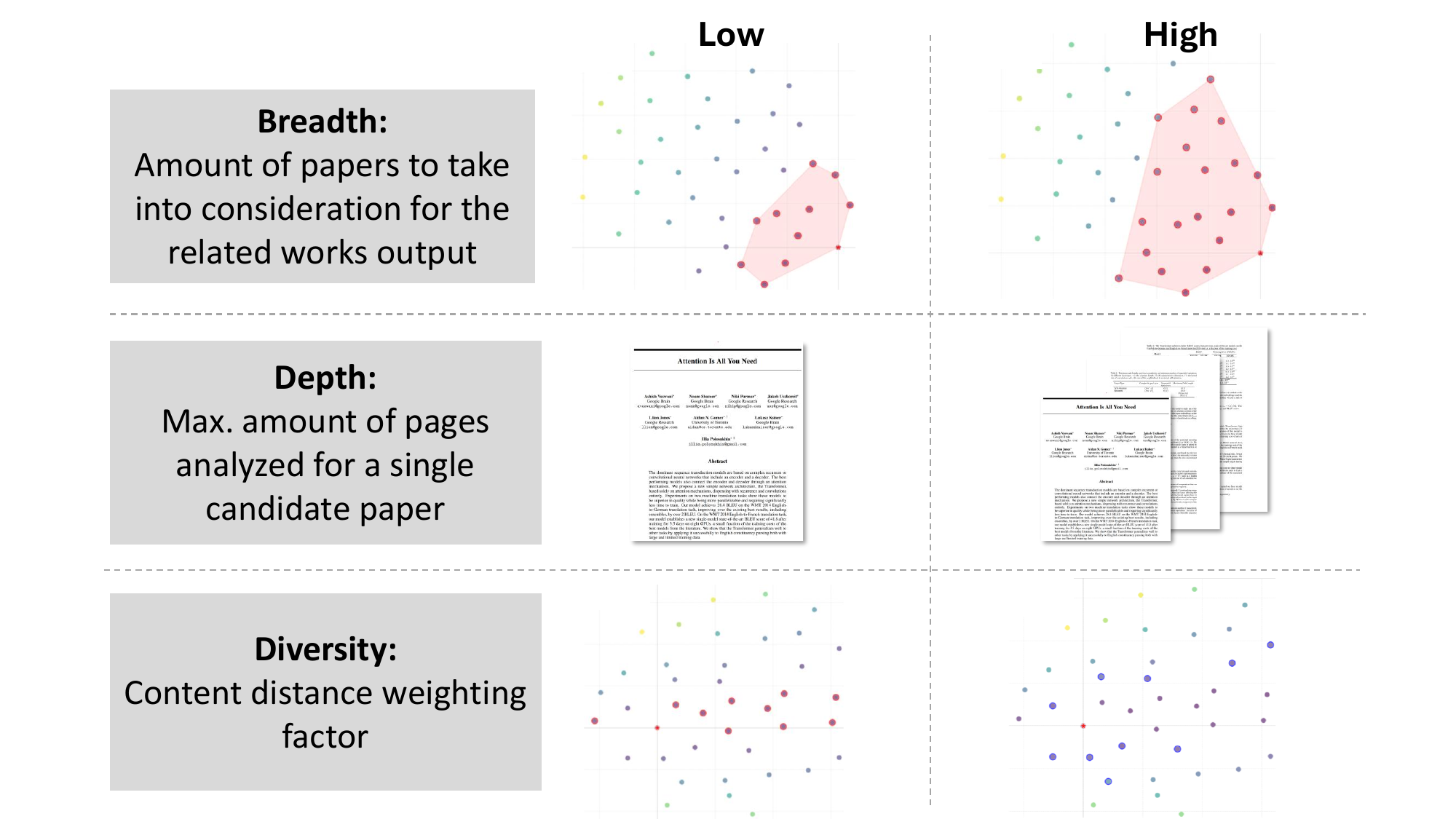}
    \caption{Visualization of the parameters \textit{breadth}, \textit{depth} and \textit{diversity} setting. Nodes represent abstract embeddings through dimensionality reduction.}
    \label{fig:Settings}
\end{figure*}

Next, we fetch the full papers for all selected candidates using their arXiv ID and extract the contained text, which is filtered using regex to exclude citations, appendix, and other irrelevant parts. We embed every page and compute similarities to the input abstract to determine a number $k$ of the most relevant pages. $k$ is another parameter, which we refer to as \textit{depth} of the generation. To choose the $k$ representative pages, we reuse the functionality from \autoref{eq:weighted-similarity}. We calculate the mean of the similarity scores and select a final set of papers, which are an aggregate of the abstract and the papers' selected pages. We refer to this set as the \textit{shortlist}. 

These papers constitute all relevant sources we employ in our generation process. To prepare the corresponding prompt, we first perform a summarization of the relevant pages and abstract of a shortlisted paper using GPT4o. We also include the source abstract in this process to tailor the resulting summary to the input paper. Finally, we aggregate the summaries in a synthesis prompt, which requests the reformulation into a joint related works section, including relevant citations, which are extracted using the arXiv library API. Finally, we filter the related works section and create a citation list that links to the relevant arXiv webpage. The full pipeline, including all steps, is displayed in \autoref{fig:pipeline}.

We consider an alternative to the candidate paper identification step if, instead of the abstract, a full (multi-page) paper is submitted by the user. In this case, we perform embedding matching over all pages and aggregate the scores. Using these, we filter for the top similarity papers, after which we fall back to the original pipeline. We note that this is only a shallow modification and we aim at investigating more sophisticated updates to the pipeline in the future.

\subsection{Update Functionality}

A primary advantage of the arXiv corpus is its frequent update schedule, which we adopt into our application. This dynamic nature is a key strength compared to static knowledge bases, particularly in the context of academic research, which is subject to constant advancements and evolving trends. However, the substantial size of the arXiv corpus requires efficient technical implementations to manage these updates effectively.

\subsubsection{Update and Reload Logic}
The system uses a hash table to manage arXiv metadata, with SHA-256 hashes computed for each entry's serialized representation. Each hash is mapped to its corresponding paper ID.
During synchronization, the system compares the hash values of incoming records with those in the hash table according to three cases:

\noindent
\textbf{(1) No Changes Detected}: When a record's hash matches the stored hash, no action is taken.

\noindent
\textbf{(2) Updates to Existing Records}: When a record's hash differs but its paper ID exists, the system recomputes the embedding and updates the database record. The hash table entry is then updated.

\noindent
\textbf{(3) New Records Detected}: For records not found in the hash table, the system computes the hash and embedding, assigns a topic using the BERTopic model, and inserts the record into the database. The hash table is updated with the new entry.

The reload process implements an iterative pipeline that processes the dataset in batches, using GPU acceleration for computing embeddings. This approach maintains synchronization between the local database and source data while avoiding redundant computation for unchanged records.

\section{Evaluation}
\label{sec:evaluation}

\begin{table*}[htbp]
    \centering
    \caption{Source relevance (Rel) and full related work quality (Qual) as evaluated by LLM-as-a-judge using GPT4o (GPT), Gemini 1.5-Pro 002 (Gem), Mistral-Large (Mist). }
    \vspace{0.2cm}
    \begin{tabular}{lccccc}
        \hline
        \textbf{Metric} & \textbf{Rel Mean} & \textbf{Rel Sum} & \textbf{Qual GPT} & \textbf{Qual Gem} & \textbf{Qual Mist} \\
        \hline
        GPT4o & 1.57 {\scriptsize ± 0.68} & 3.64 {\scriptsize ± 1.51} & 5.82 {\scriptsize ± 1.33} & 5.36 {\scriptsize ± 0.77} & 6.27 {\scriptsize ± 1.21} \\
        Agentic & 6.36 {\scriptsize ± 1.38} & 20.82 {\scriptsize ± 5.10} & \textbf{7.64} {\scriptsize ± 0.48} & 4.18 {\scriptsize ± 1.70} & 8.00  {\scriptsize ± 0.00}\\
        Citegeist & \textbf{6.90} {\scriptsize ± 0.33} & \textbf{69.00} {\scriptsize ± 3.31} & 7.55 {\scriptsize ± 0.52} & 6.91 {\scriptsize ± 0.29} & 8.27 {\scriptsize ± 0.45} \\
        Citegeist (w. paper) & 6.82 {\scriptsize ± 0.27} & 68.18 {\scriptsize ± 2.73} & \textbf{7.64} {\scriptsize ± 0.50} & 6.91 {\scriptsize ± 0.29} & \textbf{8.64} {\scriptsize ± 0.64} \\
        Source & 5.78 {\scriptsize ± 0.63} & 57.91 {\scriptsize ± 15.40} & 7.27 {\scriptsize ± 0.47} & \textbf{7.18} {\scriptsize ± 0.57} & 7.91 {\scriptsize ± 0.51} \\
        \hline
    \end{tabular}
    \label{tab:qual_gpt_std_metrics}
\end{table*}

Quantifying the quality of related work sections is difficult, as there are various relevant factors to consider. Due to resource constraints, we are unable to employ human annotators, which is why we generally fall back to employing LLM-as-a-judge. As we find that LLMs struggle with a direct comparison through exhibiting significant positional bias, we define two separate individual quality dimensions: source relevance and writing quality. For the former, we extract the source and citation abstract and prompt the model to assign a score from 0-10 depending on the relevance of the citation. For writing quality, we create a separate prompt containing the full related work section plus the source abstract and again elicit a score on a range from 0-10. Since we employ GPT4o as a judge and author in this scenario, which has been shown to lead to bias \cite{panickssery2024llmevaluatorsrecognizefavor}, we provide alternative evaluations from Mistral-Large \cite{huggingfaceMistralaiMistralLargeInstruct2407Hugging} and Gemini 1.5-Pro 002 \cite{geminiteam2024gemini15unlockingmultimodal}. We show an overview of the evaluation in \autoref{tab:qual_gpt_std_metrics}. 

For our evaluation dataset, we take 11 papers that are not yet contained in our arXiv dataset to simulate a newly written paper draft. To account for content diversity, we select five Computer Science, two Economics, one Information Science, one Physics, and two Mathematics papers. We prompt GPT4o directly to generate related work sections as a baseline. Thereafter, we manually analyze the contained citations for their correctness and replace all incorrect ones (this includes errors in author or title as well) with "(invalid citation)". Additionally, we develop a basic agentic workflow, in which we give GPT4o access to the arXiv API search functionality to perform a keyword lookup. The model is then prompted to choose the works it would like to employ based on the yielded abstracts.
For Citegeist, our application, we generate two related work sections using the input abstract and the full paper. We also evaluate the related work section from the source paper as an approximation of the gold standard. We acknowledge that the relevance metric of our approximation can be expected to be biased toward our solution since we specifically choose works based on abstract similarity. However, as we find that including full papers often does not fit into the context window, and the abstract can be considered a high-level summary of the paper, this remains a valid relevance estimation.

If not specified otherwise, all our experiments were run with \textit{breadth} ten, \textit{diversity} zero, and \textit{depth} two. We find that our solution strongly outperforms GPT4o across all metrics and is frequently rated above the source as well. Gemini tends to rate our generated works a little weaker than GPT; however, Mistral ranks our solution a little stronger, so we do not find significant implications for annotator bias. Note that we did not employ alternative annotators for the relevance since it does not contain generated text. The table also presents the aggregated relevance score sum, which we deem to be relevant since specific fields require very thorough related work sections. As expected, our base pipeline performs the best on the used relevance estimation since it only utilizes the abstract for the choice of papers. However, the solution using the full paper improves the quality estimation, hinting that this choice of works may lead to more comprehensive writing later on.

We also experiment with the effects of increasing the \textit{depth} and \textit{diversity} parameters. Setting \textit{diversity} to 0.3, we observe a small decrease in paper relevancy (-0.05) and a small increase in quality (+0.17). This is the expected outcome, and we observe that higher \textit{diversity} will often include interesting new areas but also address papers that are not directly relevant sometimes. Increasing \textit{depth} from two to six reduces relevancy (-0.31) but does not significantly change the quality (+0.01). We believe that the model struggles to focus on relevant parts once the page count exceeds a certain number during summarization, and the additional pages may dilute the overall paper relevance estimation during filtering.\newline
In addition to related work generation, we briefly explore expansion of our application to scientific question-answering and general-purpose research contextualization, which only requires prompt adaptation. While we do not conduct formal experiments, we find that the model is able to focus well on identified academic works and cites their sources correctly, resulting in appropriate answers.

\section{Conclusion}
\label{sec:conclusion}
This work presents a novel application, Citegeist, for the synthesis of scientific context retrieval and summarization on the arXiv corpus. 
We find that our pipeline provides a diverse set of functionalities and is able to generate high-quality related work analyses for various input papers. While difficult to evaluate, our experiments show clear improvement over directly prompting GPT4o and competitiveness with the source sections, even if there is a certain bias to relevance estimation. We also expose certain parameters directly to the user to influence the content of the generated section. Our pipeline is supported by a dedicated update functionality to incorporate the newest papers. We hope to address a significant gap in current research with our work and showcase strong generative capabilities based on correctly cited scientific sources. We publish our full \href{https://github.com/chenneking/citegeist}{implementation} on github and the corresponding \href{https://huggingface.co/datasets/chenneking/citegeist-milvus-db/blob/main/database.db}{dataset} on huggingface. A corresponding PyPi package is also forthcoming. A video demonstration of interaction with our website can be found on \href{https://youtu.be/gOCIdjA9FzM}{Youtube}.

\section*{Limitations}
While we are confident in the pipeline functionality, we also discovered limitations.
One drawback is the processing time for updates, with a typical two-week cycle taking approximately 4 hours without a GPU, making the process time-intensive for large-scale updates. The trade-off lies in the choice of update frequency, which may limit real-time synchronization if only done rarely. Future improvements, such as enhanced parallelization and streamlined database interactions, could help reduce processing time. 

While LLM annotators may struggle to consider this without further information, we find that some of the related works sections tend to go over works in a rather shallow manner. Increasing \textit{depth} can mitigate this to a certain degree, but after a certain point, the model struggles to consolidate all given pages. This could be addressed by adding further stages to the summarization process or employing a model with a larger context window. We also denote that we inherit the arXiv-specific content distribution, meaning, for instance, Mathematics is represented more than Philosophy. Another bottleneck is that generated sections are generally unable to accommodate more than 10-12 related works, which could potentially be solved by splitting generation across multiple prompts or, again, by choosing a model with a larger context window.

Lastly, we acknowledge that future work could expand or improve our evaluations. Since the application directly incorporates relevancy estimates, it may be considered problematic to measure this as a quality estimate. In addition, identifying related works also requires a deep understanding of several smaller aspects of the draft paper to ensure a well-rounded contextualization. Further, our study was limited by the number of draft papers we were able to gather, partially due to the manual process of verifying all cited papers of the GPT4o draft prompt. We plan to expand on this evaluation in the future, potentially by conducting an evaluation involving human annotators to see how the scoring compares.

\bibliography{main}

\appendix

\section{Web-Interface}
\begin{figure*}
    \centering
    \includegraphics[width=0.8\linewidth]{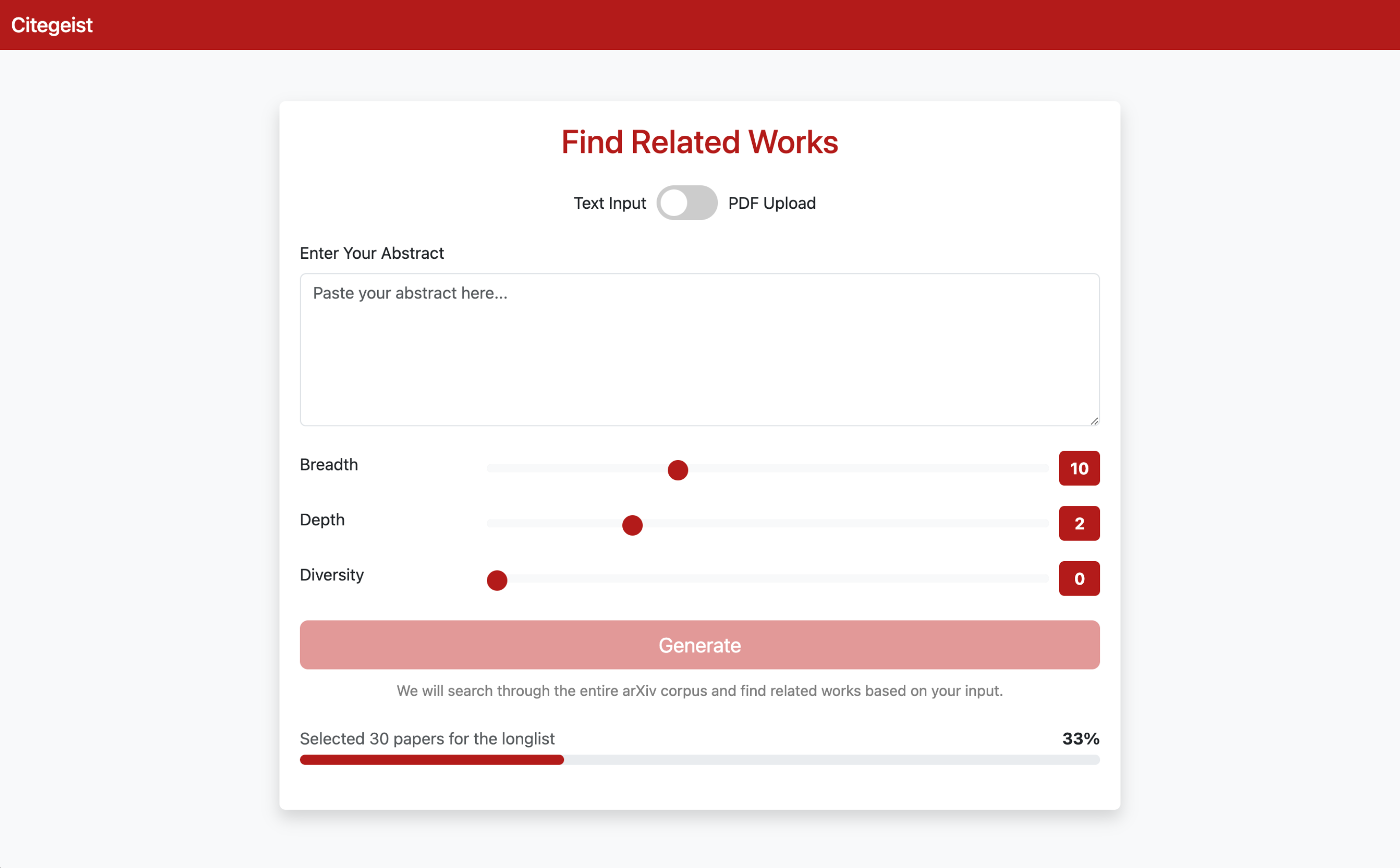}
    \caption{Citegeist Web Interface.}
    \label{fig:web-interface}
\end{figure*}

We provide access to the pipeline via a simple Web UI as shown in \autoref{fig:web-interface}. The user can provide the abstract text or upload an entire PDF document. In both cases, the user can then also tweak the three key hyperparameters used during generation.

\section{Vector Database}
A core requirement of using the citegeist pipeline is to have access to the abstract embeddings of the entire arXiv corpus. Hosting a vector database with 2.6M entries locally will likely exceed the computational capacity of some users' hardware limits and not reach the performance levels of a properly hosted and indexed vector database. Therefore, we decided to make a hosted version of this database available to all users who wish to run our pipeline. Instead of installing the database locally, they can leverage gRPC calls to our hosted version, which returns matches at low latencies. This not only improves the user experience in terms of installation but also in terms of usage, as the hosted database is many times faster than the local version. 

The database is available at \url{http://49.12.219.90:19530} and further explanations are provided in the codebase.

\section{Sample Output}
We took the abstract of this paper and ran our pipeline on it. The outputs can be found below.

\textbf{Related Work:}

In recent years, the integration of Retrieval Augmented Generation (RAG) with Large Language Models (LLMs) has emerged as a promising approach to enhance the generation of citation-backed scientific content. This approach addresses the prevalent issue of hallucinations in LLMs by incorporating retrieval mechanisms to ensure factual accuracy. Several studies have explored this domain, each contributing unique methodologies and insights that align with or complement our research on using dynamic RAG on the arXiv Corpus.

A significant body of work has focused on improving the reliability and efficiency of literature synthesis through retrieval-augmented systems. For instance, LitLLM, a toolkit designed to automate literature reviews, employs RAG principles to mitigate hallucinations in LLMs by integrating retrieval mechanisms (Agarwal et al., 2024). Similarly, OpenScholar leverages a vast datastore of open-access papers to synthesize citation-backed responses, emphasizing the importance of adapting to the continuous growth of scientific literature (Asai et al., 2024). Both studies share our goal of enhancing the accuracy and reliability of LLM-generated scientific content, highlighting the potential of retrieval-augmented systems in advancing scientific writing.

Another area of research has explored the integration of knowledge graphs and full-text grounding to improve citation accuracy and contextual relevance. The KG-CTG framework utilizes knowledge graphs to enhance citation text generation, aligning with our approach of embedding-based similarity matching and multi-stage filtering (Anand et al., 2024). Additionally, the use of Cited Text Spans (CTS) instead of abstracts for grounding citation generation further underscores the importance of full-text integration to avoid hallucinations, a challenge also addressed in our work (Li et al., 2023). These studies highlight the potential of leveraging structured data and comprehensive text sources to improve the quality of generated scientific content.

The synthesis of related work sections has also been a focal point in recent research. Shah and Barzilay's model for generating related work sections through a tree of cited papers aligns with our objectives by addressing the challenge of synthesizing information from multiple sources (Shah \& Barzilay, 2021). Similarly, the Multi-XScience dataset, designed for extreme multi-document summarization, provides a robust foundation for generating citation-backed outputs, complementing our use of the arXiv Corpus (Lu et al., 2020). These contributions emphasize the importance of content planning and summarization strategies in enhancing the coherence and relevance of related work sections.

Furthermore, innovative approaches to reference synthesis and handling have been proposed to optimize retrieval and generation processes. LLM-Ref, a writing assistant tool, enhances reference handling by retrieving and generating content directly from text paragraphs, contrasting with traditional RAG systems (Fuad \& Chen, 2024). SciLit, a pipeline for automating retrieval, summarization, and citation, employs a two-stage pre-fetching and re-ranking system to recommend papers efficiently, similar to our dynamic RAG approach (Gu \& Hahnloser, 2023). These studies underscore the need for advanced retrieval and generation techniques to improve the accuracy and contextual relevance of scientific writing.

Finally, the meta-study by Li and Ouyang on automatic related work generation highlights the challenges and lack of standardization in methodologies and evaluations, informing our efforts to enhance the accuracy and relevance of automatically generated scientific content (Li \& Ouyang, 2022). Additionally, the Retrieve-Cluster-Summarize system addresses hallucination issues by integrating document retrieval, clustering, and summarization, aligning with our emphasis on maintaining source validity while adapting to the continuous growth of the document base (Lennox et al., 2023). These insights contribute to a broader understanding of the complexities involved in generating citation-backed outputs and the potential of RAG systems in overcoming these challenges.\newline
[\textit{We have verified the correctness of all cited sources, but have not embedded them as to avoid ambiguity with our actual citations}].
\newpage
\begin{figure*}[t]
    \centering
    \includegraphics[width=\linewidth]{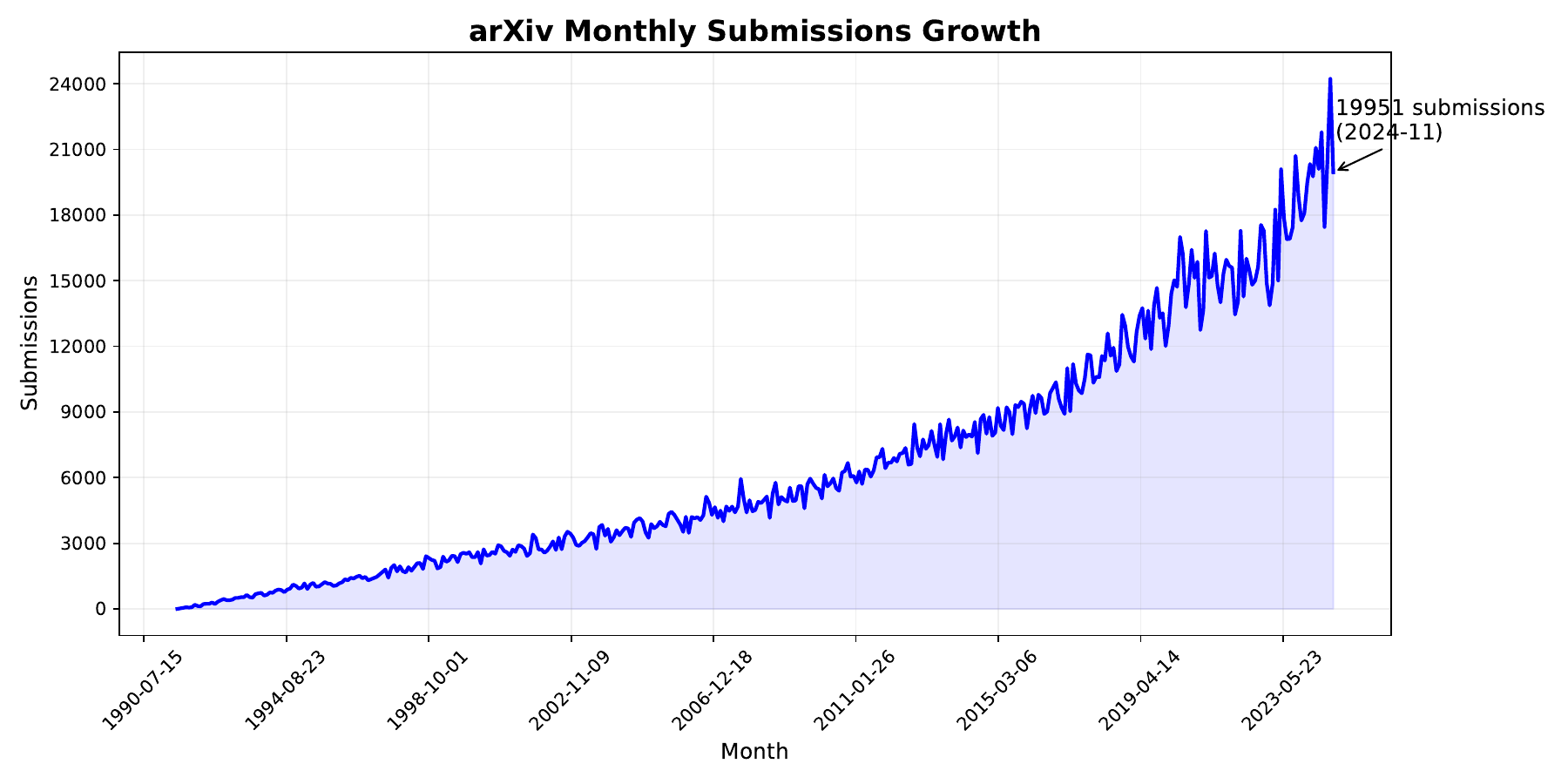}
    \caption{Number of monthly submissions to the arXiv site since establishment.}
    \label{fig:arXiv_Growth}
\end{figure*}
\begin{figure*}[t]
    \centering
    \includegraphics[width=\linewidth]{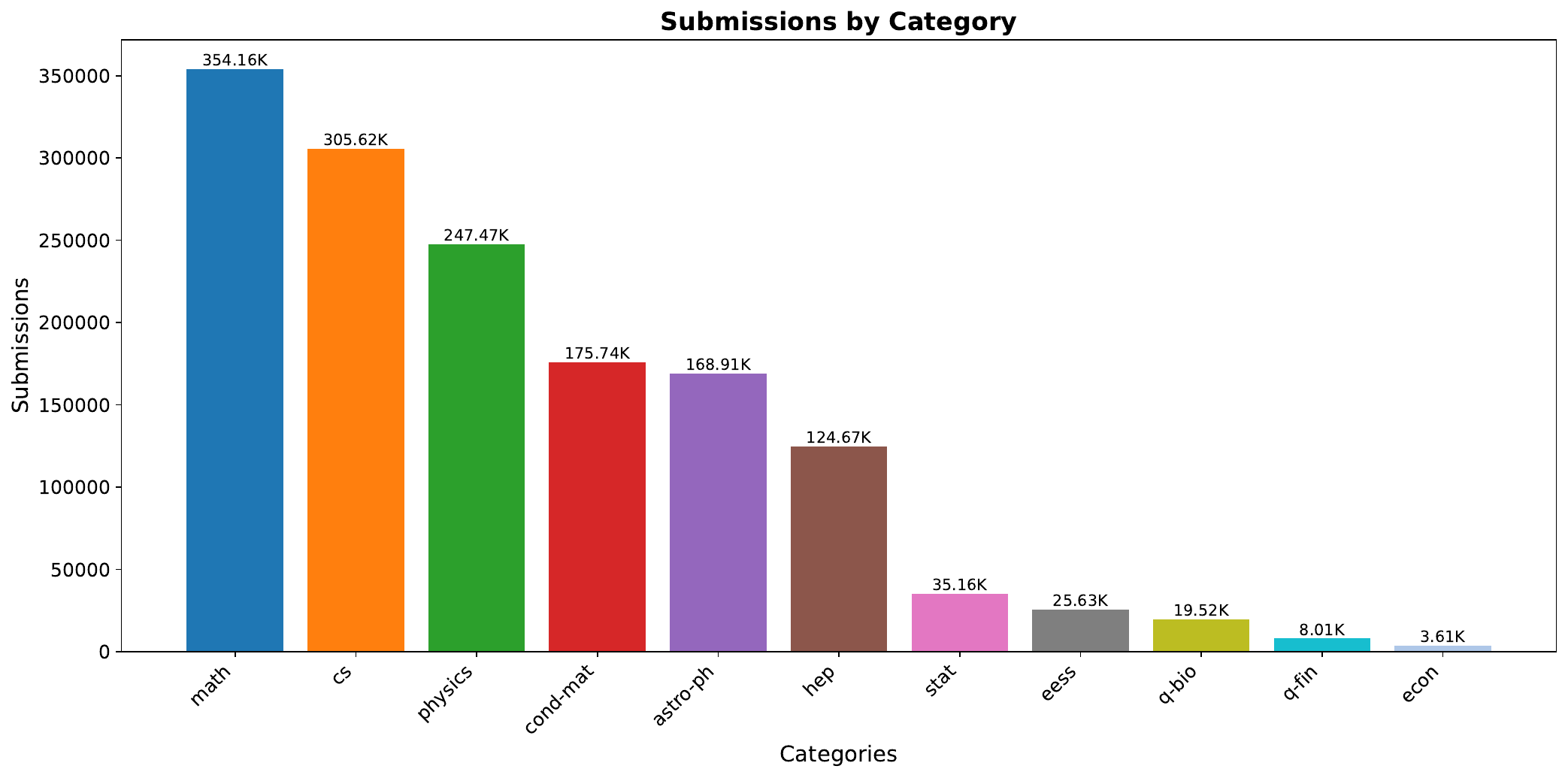}
    \caption{Histogram of topic distribution within arXiv \cite{arxiv2024categorydistribution}.}
    \label{fig:topic-histogram}
\end{figure*}

\end{document}